\documentclass[runningheads]{llncs}
\usepackage{graphicx}
\usepackage{amsmath,amssymb} 
\usepackage{color}

\newcommand{\todo}[1]{}
\renewcommand{\todo}[1]{{\color{red} TODO: {#1}}}

\usepackage[caption=false]{subfig}

\begin{document}
\pagestyle{headings}
\mainmatter
\def\ECCV18SubNumber{1735}  

\title{SConE: Siamese Constellation Embedding Descriptor for Image Matching}

\titlerunning{SConE: Siamese Constellation Embedding Descriptor}

\authorrunning{T. Trzcinski et al}

\author{Tomasz Trzcinski \inst{1, 2}\orcidID{0000-0002-1486-8906},
Jacek Komorowski \inst{1}\orcidID{0000-0001-6906-4318},
Lukasz Dabala \inst{1}, 
Konrad Czarnota \inst{1},
Grzegorz Kurzejamski \inst{1}\orcidID{0000-0002-2918-497X}, 
Simon Lynen \inst{3}}
\institute{
Warsaw University of Technology, Warsaw, Poland\\
\and
Tooploox
\and 
Google
\\
}

\maketitle

\begin{abstract}
Numerous computer vision applications rely on local feature descriptors, such as SIFT, SURF or FREAK, for image matching. Although their local character makes image matching processes more robust to occlusions, it often leads to geometrically inconsistent keypoint matches that need to be filtered out, e.g. using RANSAC. In this paper we propose a novel, more discriminative, descriptor that includes not only local feature representation, but also information about the geometric layout of neighbouring keypoints. To that end, we use a Siamese architecture that learns a low-dimensional feature embedding of keypoint constellation by maximizing the distances between non-corresponding pairs of matched image patches, while minimizing it for correct matches. The  48-dimensional floating point descriptor that we train is built on top of the state-of-the-art FREAK descriptor achieves significant performance improvement over the competitors on a challenging TUM dataset.   
\end{abstract}

\section{Introduction}
\begin{figure}[t!]
\centering
\includegraphics[width=0.40\textwidth]{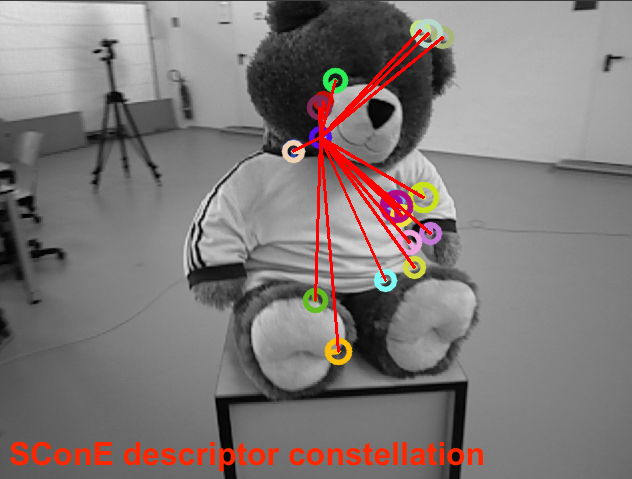}
\includegraphics[width=0.40\textwidth]{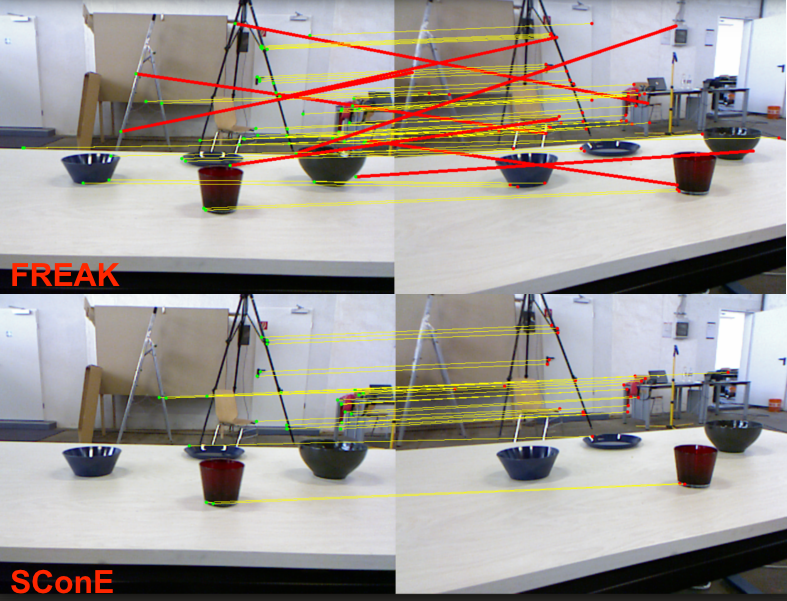}
\caption{Our proposed SConE descriptor uses information about neighbouring keypoints (left figure) to construct a discriminative low-dimensional embedding of a keypoint constellation. This way matching images with SConE reduces the number of incorrect matches found with respect to those found using a standard feature descriptor FREAK (coloured red in the top right figure) and increases the quality of resulting matches (bottom right figure).}
\label{fig:teaser}
\end{figure}

Matching images with local feature descriptors is a fundamental part of many computer vision applications, including 3D reconstruction~\cite{Agarwal2011}, panorama stitching~\cite{Brown2007} and monocular Simultaneous Localization and Mapping~\cite{Lynen15}. This topic has therefore gained significant attention from the research community~\cite{lowe2004distinctive,alahi2012freak,bay2006surf,rublee2011orb}. While traditional approaches rely on hand-crafted features~\cite{lowe2004distinctive,bay2006surf,alahi2012freak,rublee2011orb}, more recent descriptors use machine learning techniques such as boosting~\cite{Trzcinski13a} or deep learning~\cite{Simo_iccv2015,yi2016lift} to train discriminative transformation-invariant representations. Although using local feature descriptors proposed in the literature increases robustness of image matching methods to partial occlusions, it often leads to incorrect descriptor matches, as presented in the upper right part of Fig.~\ref{fig:teaser}. 
In this paper, we propose a more discriminative feature descriptor by encoding information about constellation of keypoints, as shown in Fig.~\ref{fig:teaser}.
To that end, we use a Siamese neural network that learns low-dimensional feature embeddings by minimizing distance between similar keypoint constellations, while maximizing it for non-matching pairs. Instead of relying on a local intensity patch around the detected keypoint, we construct our embedding by feeding into the neural network information on a central keypoint and its nearest neighbourhood keypoints on the image.  
The resulting 48-dimensional Siamese Constellation Embedding descriptor, dubbed SConE for simplicity, is built using FREAK~\cite{alahi2012freak} as a base descriptor, however our framework is agnostic to descriptor types and can be generalized to other descriptors. 
Evaluation of our descriptor on the challenging TUM dataset~\cite{sturm2012benchmark} shows that despite its compact nature, SConE outperforms its competitors, while decreasing the computational cost of matching by eliminating the need for a geometrical verification step.

In the remainder of this paper, we first discuss related work in Section~\ref{sec:related}. We then describe the details of our method in Section~\ref{sec:method}. Finally, in Section~\ref{sec:evaluation} we show that our descriptor is able to outperform the state-of-the-art descriptors on a real-life dataset and we conclude the paper in Section~\ref{sec:conclusions}.

\section{Related Work}
\label{sec:related}

Due to the role of local features descriptors in many computer vision tasks, significant amount of work has been focused on building those representations effectively and efficiently~\cite{lowe2004distinctive,bay2006surf,alahi2012freak,Simo_iccv2015,Trzcinski13a}. Floating-point descriptors, such as SIFT~\cite{lowe2004distinctive} or SURF~\cite{bay2006surf}, typically offer better performance at a higher computational cost. Their binary competitors, such as FREAK~\cite{alahi2012freak} or ORB~\cite{rublee2011orb} approximate many operations and simplify the resulting representation to a binary output. Our proposed method is built on top of the binary FREAK descriptor, which offers an efficient yet powerful alternative to floating-point competitors. Nevertheless, the framework proposed in this paper is general enough to be applicable also to other descriptors.

Recently, due to their success in other domains, deep neural networks have also been used to train feature descriptors~\cite{yi2016lift,Simo_iccv2015}. For instance, LIFT~\cite{yi2016lift} uses a neural network architecture to handle full pipeline of feature extraction and computation.  Another method called MatchNet uses Siamese neural network to jointly learn feature representation and matching procedure~\cite{han2015matchnet}. In~\cite{loquercio2017efficient}, they use a triplet loss function coupled with convolutional neural network that aims at training context-augmented descriptors based on FREAK. 
In our work, we use a Siamese architecture to learn low-dimensional feature embeddings based on FREAK descriptors. But instead of using an image patch around detected keypoints, we incorporate data on the neighbourhood keypoints.

\cite{Niepert:2016:LCN:3045390.3045603}
\cite{Defferrard:2016:CNN:3157382.3157527}
 apply convolutional neural networks to graph data to learn useful features.
However our input data does not have a graph structure defined by an adjacency matrix.
Spatial positions of neighbourhood keypoints and its attributes (binary descriptor, position and orientation) are important, not the structural relationships between keypoints.

Once keypoint descriptors are extracted, they are typically matched with each other to find correspondences between image regions. Depending on the final application, the matching can be done using brute-force or approximate nearest neighbour (ANN) search methods~\cite{muja2012fast},~\cite{muja2014scalable}. Heuristic techniques (e.g. two nearest neighbour ratio test~\cite{lowe2004distinctive}) are used to filter out outliers. In the final stage, putative matches are typically subject to geometric validation using epipolar constraint with a robust parameter estimation method, such as RANSAC~\cite{fischler1981random} or its extension USAC~\cite{raguram2013usac}. 
A recently proposed approach called GMS (Grid-based Motion Statistics)~\cite{bian2017gms} also aims at filtering out geometrically incorrect matches using a simple heuristic based on the number of matches in the keypoint neighbourhood. Although often effective, above methods require additional computational cost.
In our method we propose to embed the geometrical information useful for filtering the matches within the descriptor itself. This way we can avoid the unnecessary post-processing step and increase the efficiency of the image matching pipeline.

\section{Method}
\label{sec:method}

Our method aims at improving precision of the descriptor matching step. 
Instead of matching raw descriptors (e.g. 512-bit FREAK descriptors), we compare more discriminative representations of keypoint constellations.
We define a \emph{constellation} as a set of nearby keypoints in an image. It consists of a \emph{central keypoint} and its $k$ nearest, in Euclidean distance sense, keypoints detected on the same image. 
An exemplary constellation is visualized on Fig. \ref{fig:teaser}
The following information is taken into account when constructing a \emph{constellation}: binary descriptor, scale and orientation of a central keypoint; and binary descriptors, relative position (with respect to the central keypoint), scale and orientation of each of its $k$ nearest neighbours.
In this work, based on an initial experiments, we set $k=20$.

\begin{figure}
\centering
\includegraphics[width=0.6\textwidth]{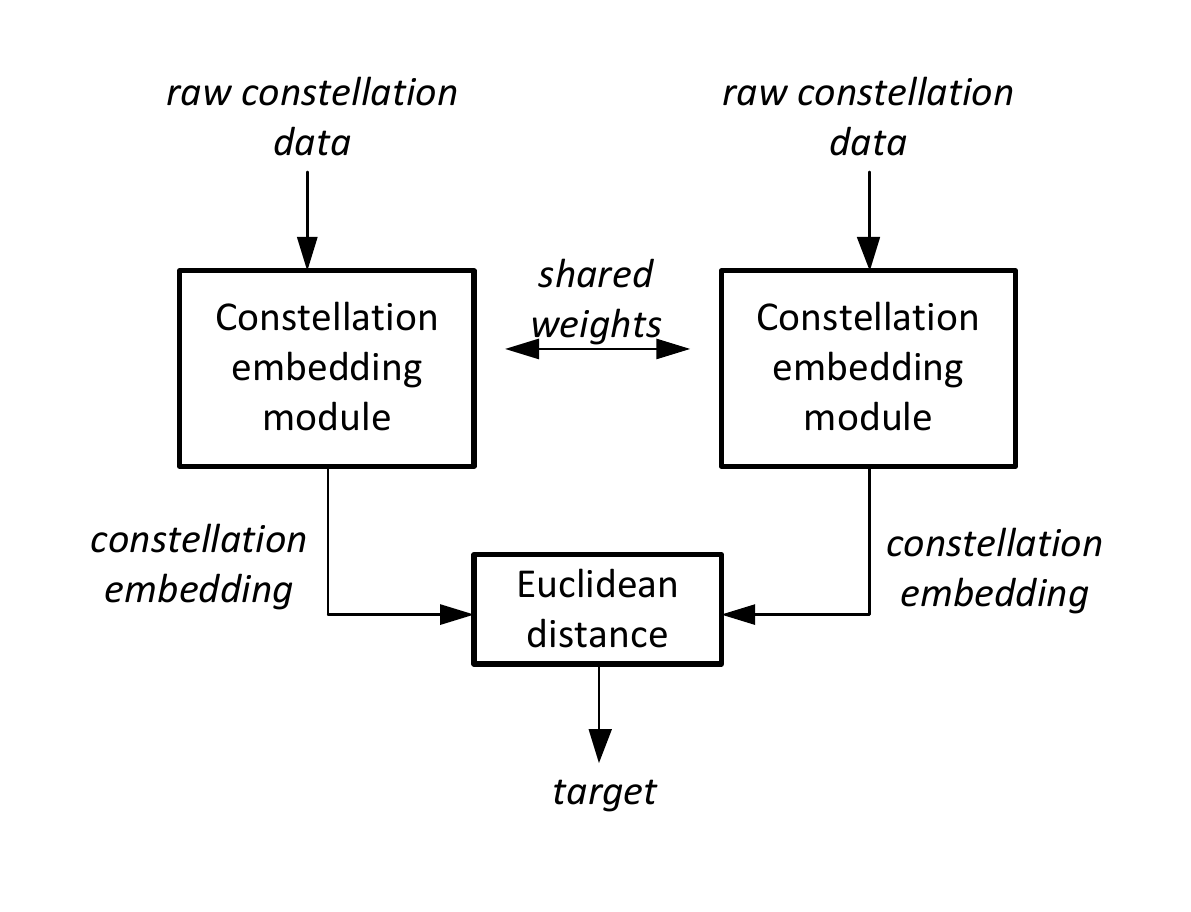} \quad
\includegraphics[width=0.85\textwidth]{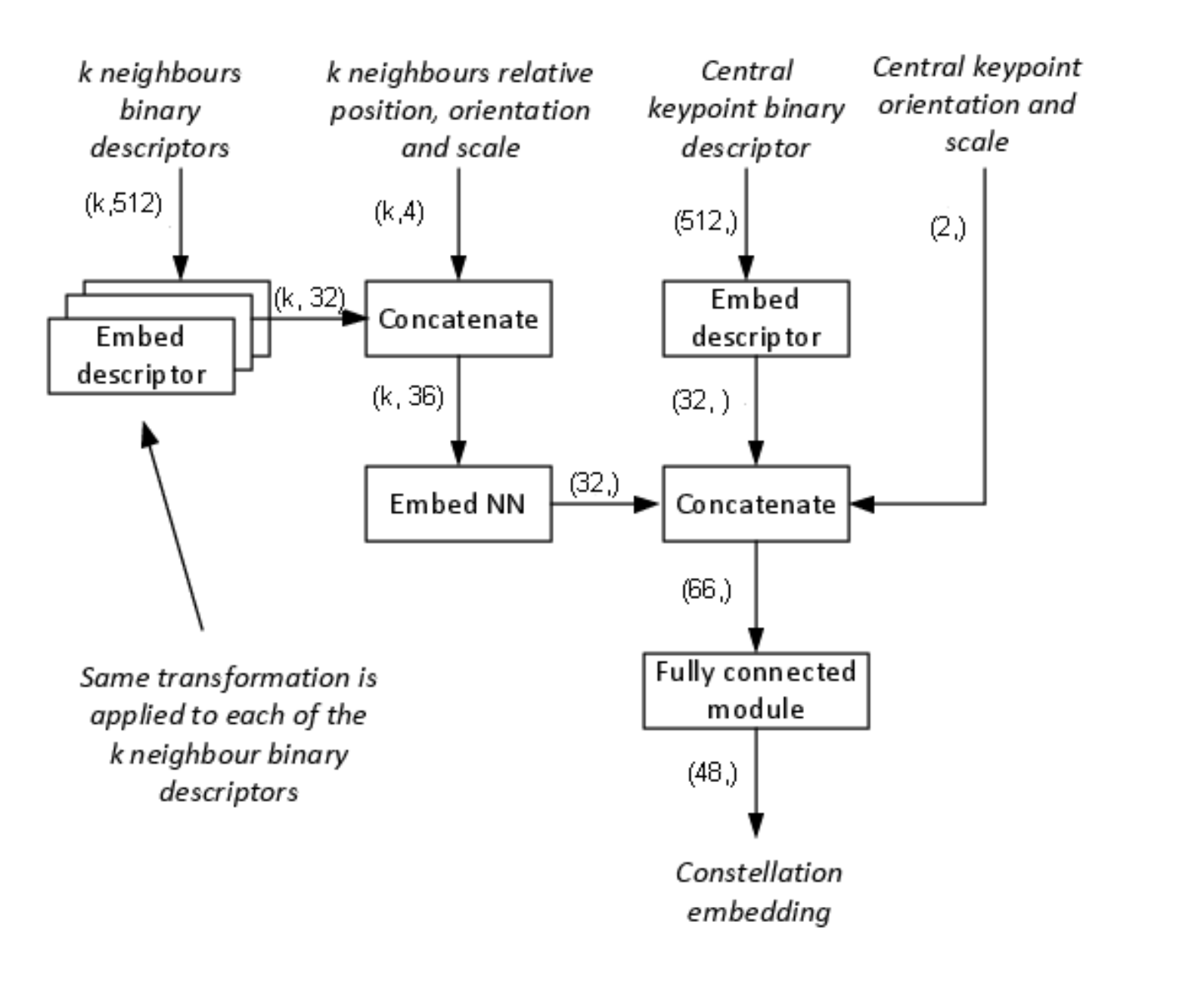}
\caption{(Top) High level architecture of the Siamese neural network computing constellation embeddings. (Bottom) Architecture of a constellation embedding module of a Siamese neural network. 
We feed into the network information on central keypoint and its nearest, in Euclidean distance sense, neighbourhood keypoints on the image.  Central keypoint FREAK descriptor, orientation and scale are used along with nearest neighbours, in Euclidean distance sense, informations. That include FREAK descriptors, their relative positions, orientations and scales with respect to the central keypoint.
Hence, our resulting SConE descriptor of the central keypoint offers better performance of descriptor matching at a lower computational cost than competing descriptors.}
\label{fig:siamese2}
\end{figure}

Dimensionality of the data constituting a constellation is rather high. 
We find low dimensional constellation embeddings by training the Siamese neural network~\cite{Bromley:1993:SVU:2987189.2987282}. This produces low dimensional, real valued, embeddings that can be efficiently stored and processed.
High-level architecture of our Siamese neural network is depicted on Fig. \ref{fig:siamese2}. The network consists of two identical Siamese modules (same network architecture with shared weights) that compute low dimensional constellation embeddings. Representations computed by Siamese modules can be matched using standard Euclidean distance between them.
 

The network is trained by presenting mini-batches consisting of pairs of similar and dissimilar constellations. 
We consider two constellations similar if their central keypoint is a projection of the same 3D scene point (landmark). Otherwise constellations are dissimilar.
We use a contrastive loss function, as formulated in \cite{Hadsell:2006:DRL:1153171.1153654}. 
Let \( X_1, X_2 \) be a pair of constellations in the training set and $Y$ a binary label assigned to this pair. $Y = 0$ if constellations $X_1$ and $X_2$ are similar, and  $Y = 1$ if they are dissimilar. 
\( D_W \) is a parametrized distance function between constellations $X_1$ and $X_2$, defined as an Euclidean distance between learned constellation embeddings \( G_W \).
\begin{equation}
D_W \left( X_1, X_2 \right) = || G_W \left( X_1 \right) - G_W \left( X_2 \right) ||_2
\end{equation}
The loss \( \mathcal{L} \) function minimized during the training is defined as:
\begin{equation}
\mathcal{L} = \sum_{i=1}^{P} L \left( W, \left( Y, X_1, X_2  \right) ^i \right),
\end{equation}
with
\begin{equation}
L \left( W, \left( Y, X_1, X_2  \right) ^i \right) 
=
\left( 1-Y \right) L_S \left( D^i_W \right)
+
Y
L_D \left( D^i_W \right)
\end{equation}
where \( \left( Y, X_1, X_2 \right) ^i \) is the i-th sample composed of a pair of constellations  \(X_1, X_2 \) and a binary label.
\( L_s = \frac{1}{2} \left( D_W \right) ^2 \) is a partial loss function for a pair of similar constellations and
\( L_D = \frac{1}{2} \left( \max \left\lbrace 0, margin - D_W \right\rbrace \right) ^2 \) is a partial loss function for a pair of dissimilar constellations.

As shown in Fig.~\ref{fig:siamese2}, a constellation is fed to the \emph{constellation embedding module} as a high dimensional vector. We process binary descriptor (512 dimensions for FREAK), scale and orientation (2 dimensions) of the central descriptor. For each neighbourhood keypoint, we use its binary feature descriptor (512 dimensions) and relative position, scale and orientation (4 dimensions).
For a constellation consisting of the central descriptor and its 20 nearest neighbours, this gives $514 + 20 \times 516 = 10 834$ dimensions.
We designed the \emph{constellation embedding module} (see bottom of Fig.~\ref{fig:siamese2}) in a modular fashion. The design was based on an extensive series of experiments to help us identify the best architecture of each component. The best performing architecture is described below.
The twin \emph{constellation embedding module} first computes $k$ 32-dimensional embeddings of $k$ neighbour binary descriptors. The resulting embeddings are concatenated with $k$ neighbours relative position, relative orientation and relative scale with respect to the central keypoint. This gives a sequence of $k$ 36-dimensional vectors which are further processed by RNN (recurrent neural network) module producing 32-dimensional neighbourhood representation.
Then neighbourhood representation is concatenated with the central keypoint binary descriptor, orientation and scale resulting in 66-dimensional vector. This is processed by a final fully connected module resulting in the 48-dimensional constellation embedding called SConE for Siamese Constellation Embedding descriptor. 
Details of each component are given in Table \ref{lab:siamese3}.
The size of the final embedding (48 real values) was chosen as a compromise between the descriptor discriminative power and the storage requirements.

\begin{table}[t]
\begin{center}
\caption{Components of a twin constellation embedding module from Fig. \ref{fig:siamese2}. Scaled Exponential Liner Unit (SELU) \cite{NIPS2017_6698} activation is used after each fully-connected layer. }
\label{lab:siamese3}
\begin{tabular}{l p{8cm}}
\hline\noalign{\smallskip}
{\bf Component name}			& {\bf Details}   \\ 
\noalign{\smallskip}
\hline
\noalign{\smallskip}
Embed descriptor      &  3 fully-connected layers with 512/256/32 units
\\ 
\noalign{\smallskip}
Embed NN & 2 layer bidirectional LSTM \cite{hochreiter1997long} with 32/32 units followed by 3 fully-connected layers with 64/64/32 units
\\ 
\noalign{\smallskip}
Fully-connected module   & 3 fully-connected layers with 64/64/48 units \\ 
\noalign{\smallskip}
\hline
\end{tabular}
\end{center}
\end{table}

Siamese neural network training is conducted using data acquired with structure-from-motion solution embedded in a Google Tango tablet. The device produces datasets containing keypoints and feature descriptors detected on the recorded video sequences. Camera poses and scene structure, in the form of sparse 3D point sets, are reconstructed using reliable structure-from-motion techniques.
The training set was constructed by concatenating samples from multiple video sequences.
It consists of almost 10 thousand keyframes with over 4 million FREAK descriptors linked with 259 thousand landmarks.
The validation sequence, used to measure the performance of the trained networks, contains almost 5 thousand keyframes with over 2 million feature descriptors linked with 120 thousand landmarks.
In both sequences, almost half of the feature descriptors are linked with reconstructed 3D scene points (landmarks) whereas the rest of them is not linked with any landmark.

\begin{figure}[t]
\centering
\includegraphics[width=0.49\textwidth]{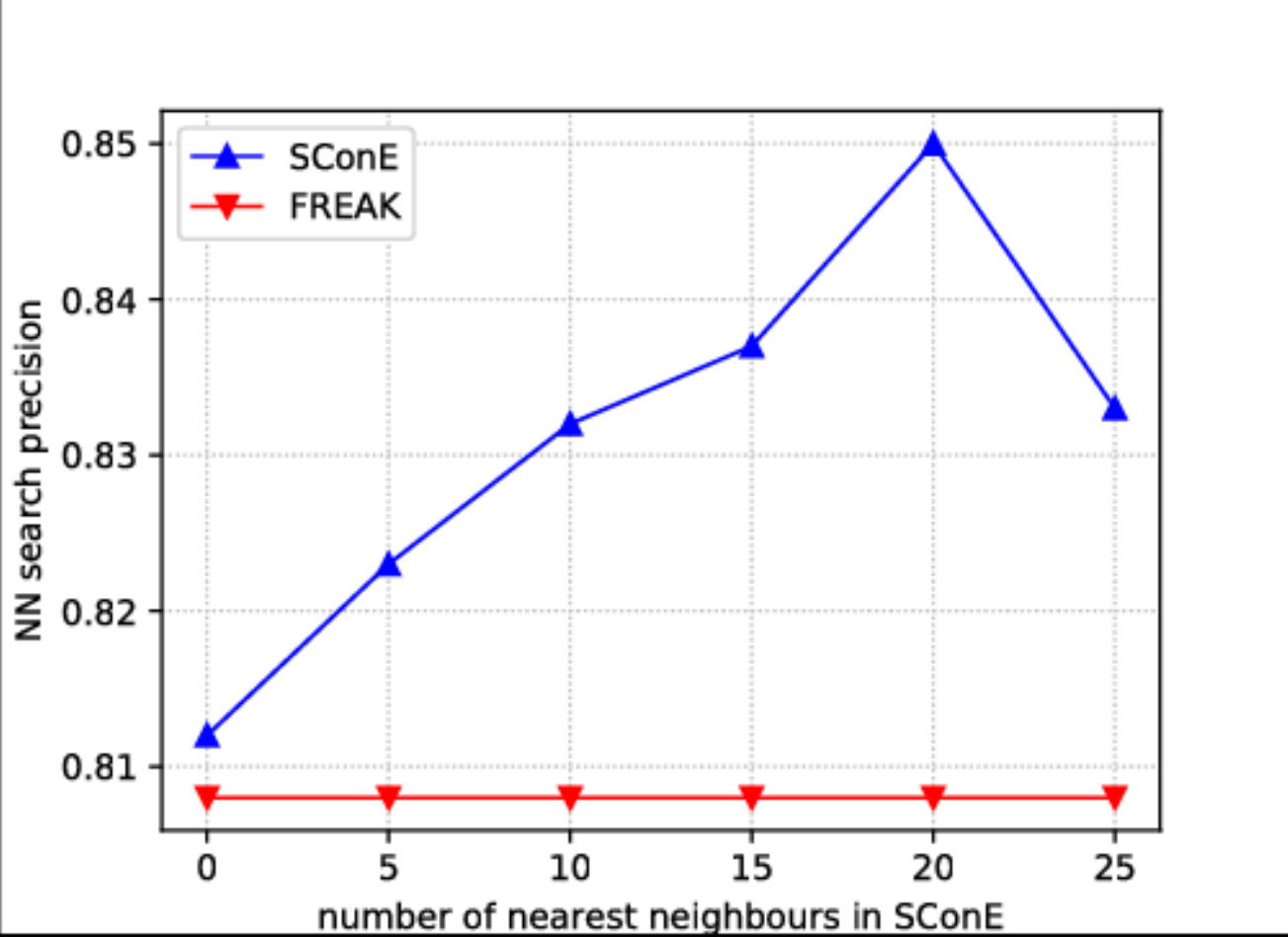}
\caption{The influence of k nearest neighbours on the nearest neighbour search precision in the validation set. The best results are achieved when 20 neighbourhood keypoints are used to form a constellation.}
\label{fig:knn_results}
\end{figure}

We experimentally choose the number of neighbours used to form the constellation, that produces the most discriminative SConE descriptor.
This is done by training the Siamese network multiple times using constellations of various size and evaluating the performance of the trained network.
We use nearest neighbour search precision on the embeddings of the validation set as the performance measure. 
The precision is calculated as follows. First, embeddings of validation set elements are calculated using the constellation embedding module of the trained Siamese network.
Then 10 thousand embeddings is randomly chosen from the validation set. For each sampled embedding, its nearest  neighbour in embedding space (that is in Euclidean space, as embeddings are real-valued vectors) is found. 
If the nearest neighbour is linked with the same 3D scene point (landmark) as the sampled element we declare a match. \emph{Precision} is calculated as the percentage of correct matches.
The results are depicted on Fig.~\ref{fig:knn_results}. As the number of neighbours increases the precision grows, to reach a maximum for 20 neighbours. Compared to using raw FREAK descriptors we get increase of nearest neighbour search precision from 0.807 to 0.851 on our validation set.
 
\section{Evaluation}
\label{sec:evaluation}

This section describes evaluation procedure and its results. The evaluation dataset is described in Section~\ref{sec:dataset}. 
In Section~\ref{sec:evaluation_procedure} we present our evaluation protocol. 
Finally, in Section~\ref{sec:results} we show the results of our evaluations.

\subsection{Dataset}
\label{sec:dataset}

\begin{figure}[!t]
\centering
\includegraphics[width=\columnwidth]{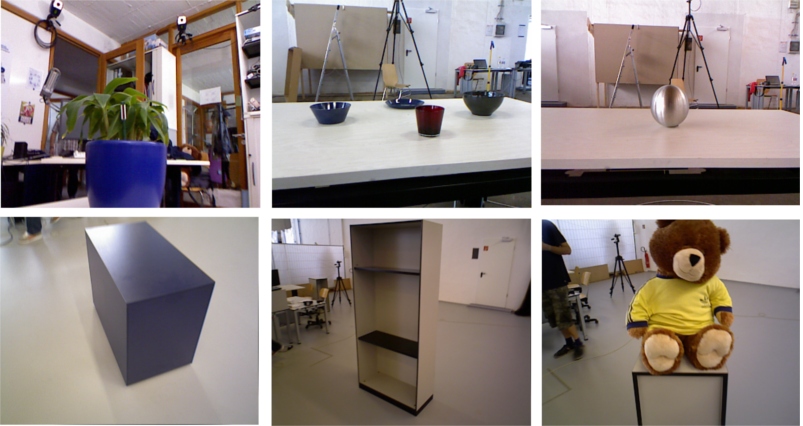}
\includegraphics[width=\columnwidth]{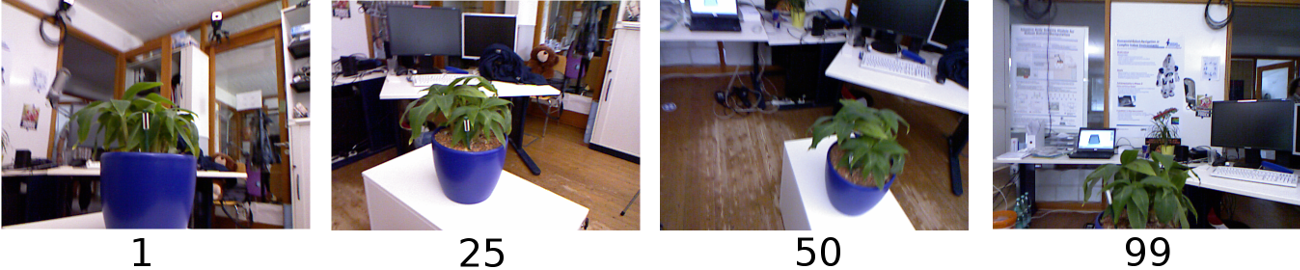}
\caption{(Top) Exemplary images from TUM~\cite{sturm2012benchmark} video sequences used in evaluation of our method.
(Bottom) 1st, 25th, 50th and 99th keyframes from one test sequence ({\tt fr1/plant}). There's a large viewpoint variation in frames forming one sequence.}
\label{fig:frames}
\end{figure}

For the evaluation procedure, we use a challenging TUM dataset~\cite{sturm2012benchmark}, often used in other works to compare descriptors' performance~\cite{bian2017gms}. TUM is a large dataset with sequences recorded using Microsoft Kinect sensor and we choose seven of them for evaluation: 
{\tt fr1/plant}, 
{\tt fr2/dishes}, 
{\tt fr2/metallic\_sphere2}, 
{\tt ft3/cabinet}, 
{\tt fr3/large\_cabinet},
{\tt fr2/flowerbouquet}
and {\tt fr3/teddy}. 
Sample sequences can be seen in top of Fig.~\ref{fig:frames}. In addition to images, sensor ground-truth trajectory and depth-maps are provided. We divide each sequence into 100 long subsequences. First frame is treated as a reference and 99 others are used for matching. Bottom of Fig.~\ref{fig:frames} presents four different frames from a sequence. They differ significantly, the last one being rotated almost 360 degrees. It makes it hard for matchers to find any correct matches between first and last frame in such scenario.

Due to a lot of blurred images and changes in lighting, the TUM dataset is considered to be rather challenging. Additional difficulty comes from the fact that color images, depth maps and camera positions are not perfectly consistent. They were collected in different moments of time, so timestamps cannot be perfectly aligned and to address this problem we approximate them to minimize the time gaps between frames. 
Furthermore, due to the limitation of the capturing device, a large portion of depth maps does not provide correct depth values, especially on the edges of objects where a large portion of keypoints is detected. 
Hence we use an epipolar geometry condition and ground truth camera poses to verify correctness of a match.

\begin{figure}
\centering
\includegraphics[width=0.49\textwidth]{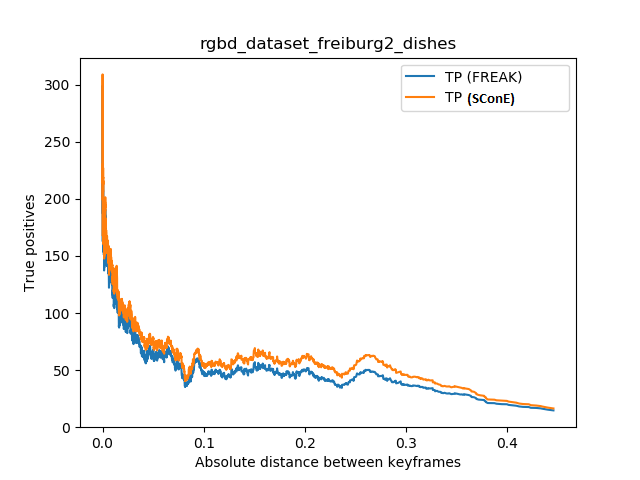}   \includegraphics[width=0.49\textwidth]{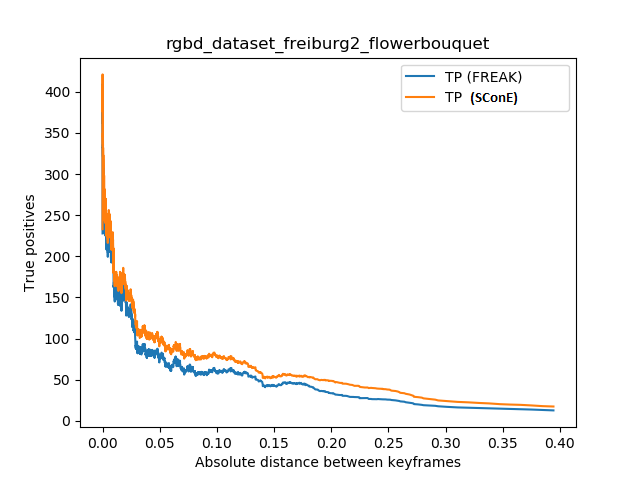}
\caption{Number of true positives in matching feature descriptors between a pair of images as a function of an angular distance  between keyframes. Results on {\tt fr2/dishes} (left) {\tt fr2/flowerbouquet} (right) sequences from TUM~\cite{sturm2012benchmark} dataset are presented.
SConE consistently yields better results than FREAK descriptor due to encoding additional information about the neighbourhood keypoints.}
\label{fig:gt1}
\end{figure}

\begin{figure}
\centering
\includegraphics[width=0.49\textwidth]{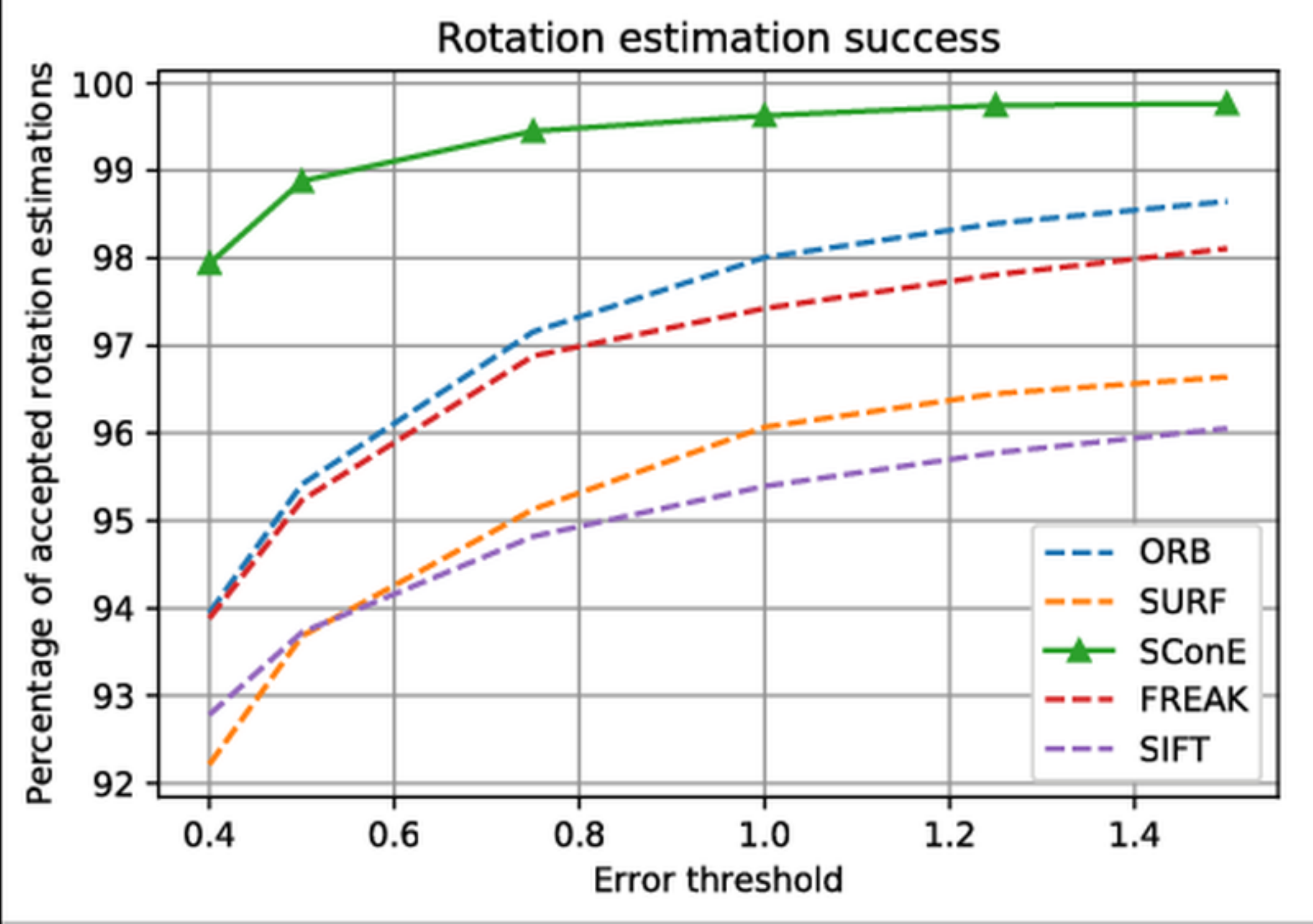}
\includegraphics[width=0.49\textwidth]{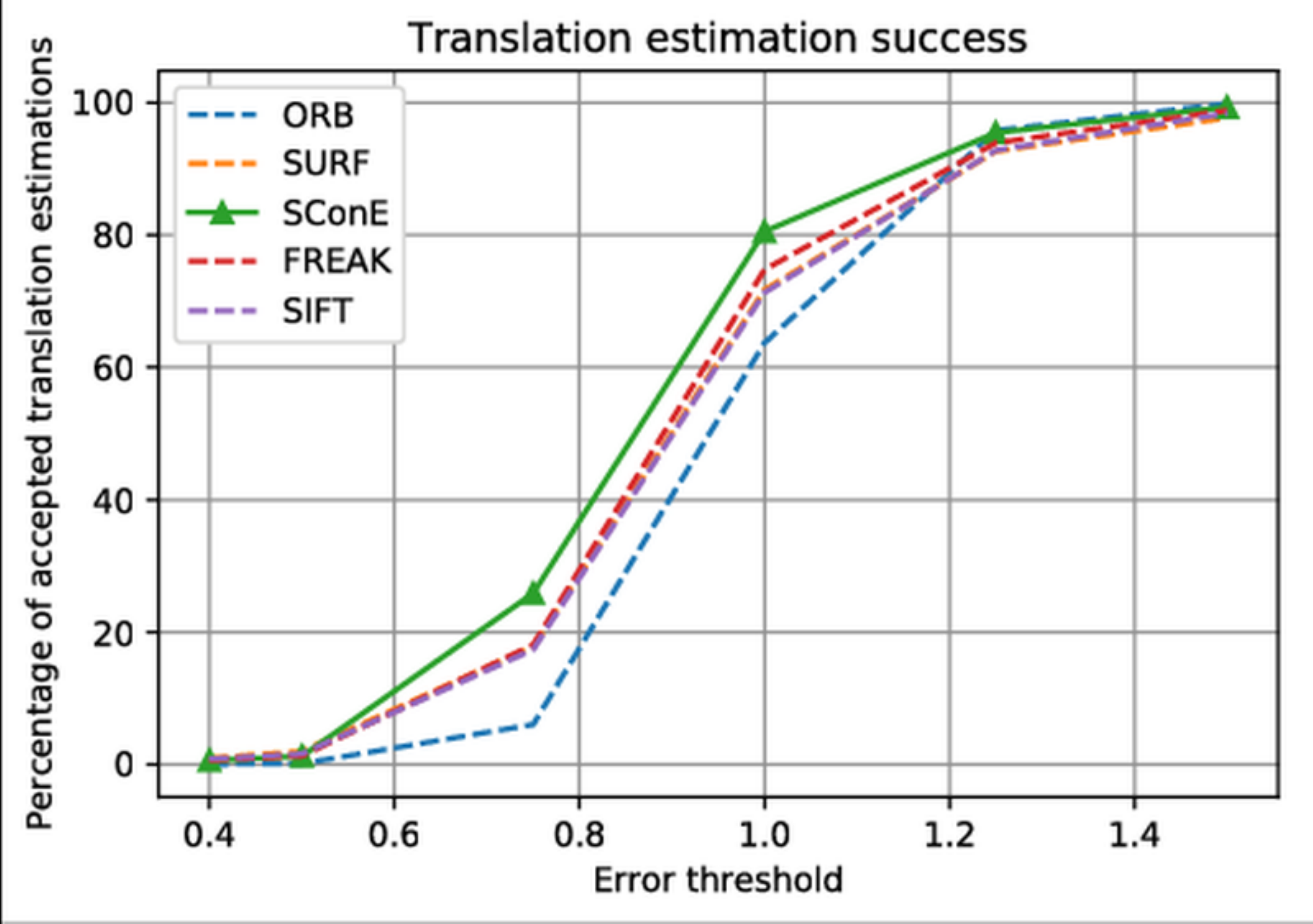}
\caption{Pose estimation errors on TUM~\cite{sturm2012benchmark} dataset for SConE and competing descriptors. SConE outperforms the state-of-the-art descriptors, including SIFT and SURF, across all error thresholds.}
\label{fig:sfm_results2}
\end{figure}

\subsection{Evaluation procedure}
\label{sec:evaluation_procedure}

We test our descriptor in a demanding scenario of a real application, strictly connected with SLAM and Structure from Motion pipelines. We use TUM's ground truth presented as a trajectory and calibration data for each camera in the set. The TUM dataset is specifically designed to evaluate Structure from Motion algorithms and therefore its frame resolution is low and graphical content is often lacking the details necessary to track dense feature sets. Nevertheless, such characteristics create a demanding benchmark for camera pose estimation and we therefore use it in our evaluation. 

We compare our method to the state-of-the-art methods for image matching: FREAK \cite{alahi2012freak}, SURF \cite{bay2006surf}, SIFT \cite{lowe2004distinctive}, ORB \cite{rublee2011orb} (implementations comes from OpenCV \cite{OpenCV} package), GMS \cite{bian2017gms} and embeddings calculated by our custom artificial neural network (SConE). 
For SConE, we first use  FAST \cite{rosten2006machine} key point detection algorithm. Then we compute FREAK feature descriptors at detected keypoints. 
SConE descriptors are calculated using the constellation embedding module of the trained Siamese network. The network training is performed as described in Section \ref{sec:method}.
For each 100 frame subsequence from the evaluated TUM sequence, we compute matches between the first frame and all others, resulting in 99 image pair matches. For each image pair we find pairs of corresponding features  with a brute-force approach. 
Matches are then filtered using standard ratio-test.

We compute essential matrix from the key point correspondences for each image pair. From them we estimate relative camera pose for each pair of images in form of a rotation matrix and translation vector. We use the OpenCV \cite{OpenCV} implementation with RANSAC for this purpose. 

RANSAC is needed in the process because of two reasons. The first is its ability to filter out matches considered good given a 3D model, but giving perturbations in affine transformation estimation. This situation happens when the scene contains moving or deforming objects. The second factor is connected with a level of locality in SConE. SConE makes use of constellation, incorporating structural data of a bigger area than the base descriptor itself, but still is considered as a local descriptor. If duplicate elements of the scene appear, SConE is prone to generate bad matches. Feature duplicates can be seen in various real case scenarios where textures contain patterns or multiple features of the same appearance, for instance in windows or buildings' facades.

We compare the relative pose recovered using abovementioned procedure against ground truth and calculate qualitative metric for each image pair. The metric is presented as an error in translation and rotation estimation, calculated according to the procedure described in the Odometry Development Kit from KITTI benchmark \cite{Geiger2012kitty}. KITTI benchmark defines a method of error calculation for 3D tracks with six degrees of freedom with asynchronous sampling. In our case the data is synchronized, so the formulas are straightforward. Error in translation is calculated as a translation vector difference in 3D. Error for rotation is calculated from relative 3x3 rotation matrix \emph{dR} according to the formula:

\begin{equation}
d = \frac{tr\left(\emph{dR}\right) - 1.0}{2}
\end{equation}
\begin{equation}
R_{err} = acos\left(\max\left(\min\left(d, 1.0\right), -1.0\right)\right)
\end{equation}

\begin{figure}
\centering
\includegraphics[width=0.49\textwidth]{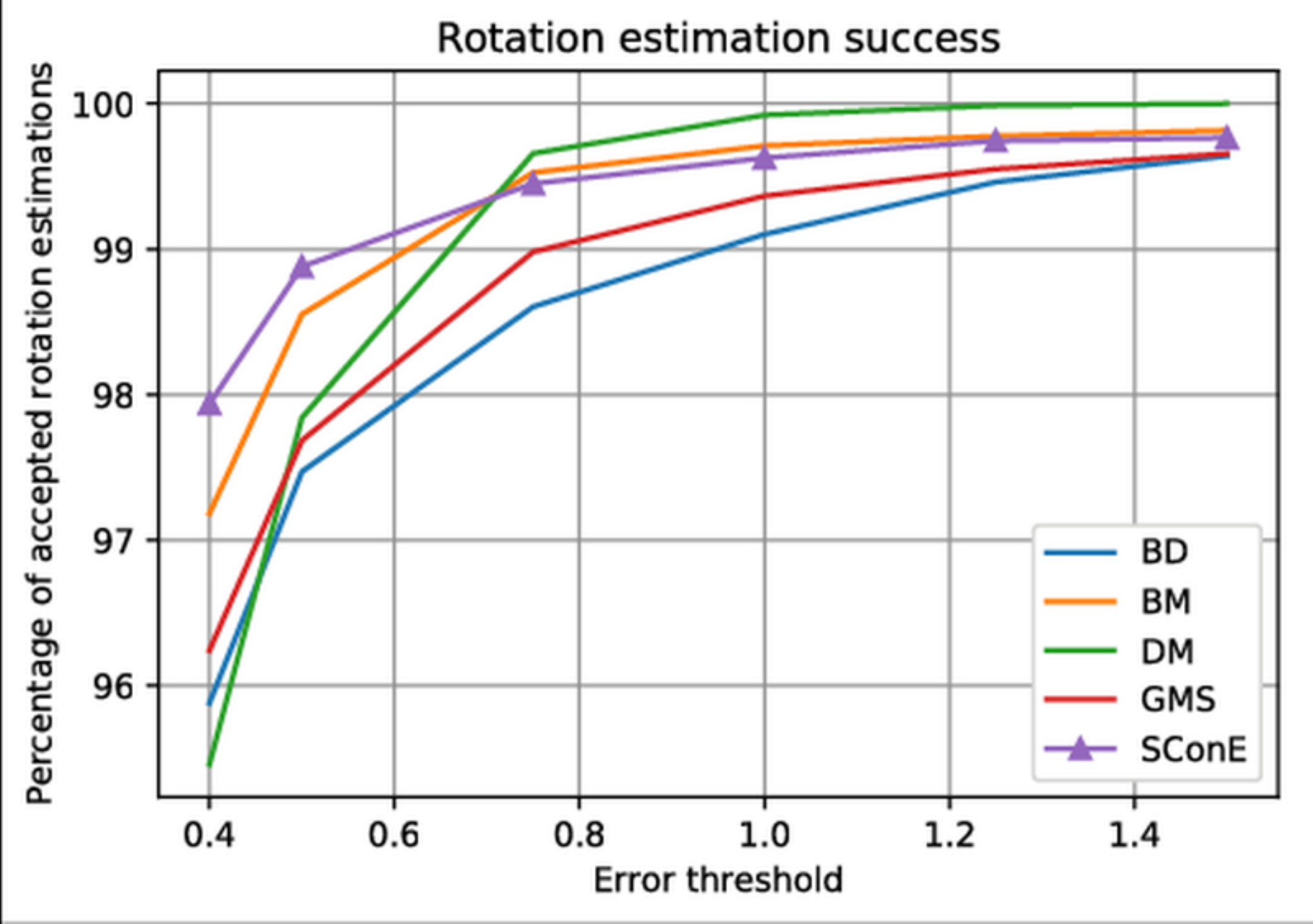}
\includegraphics[width=0.49\textwidth]{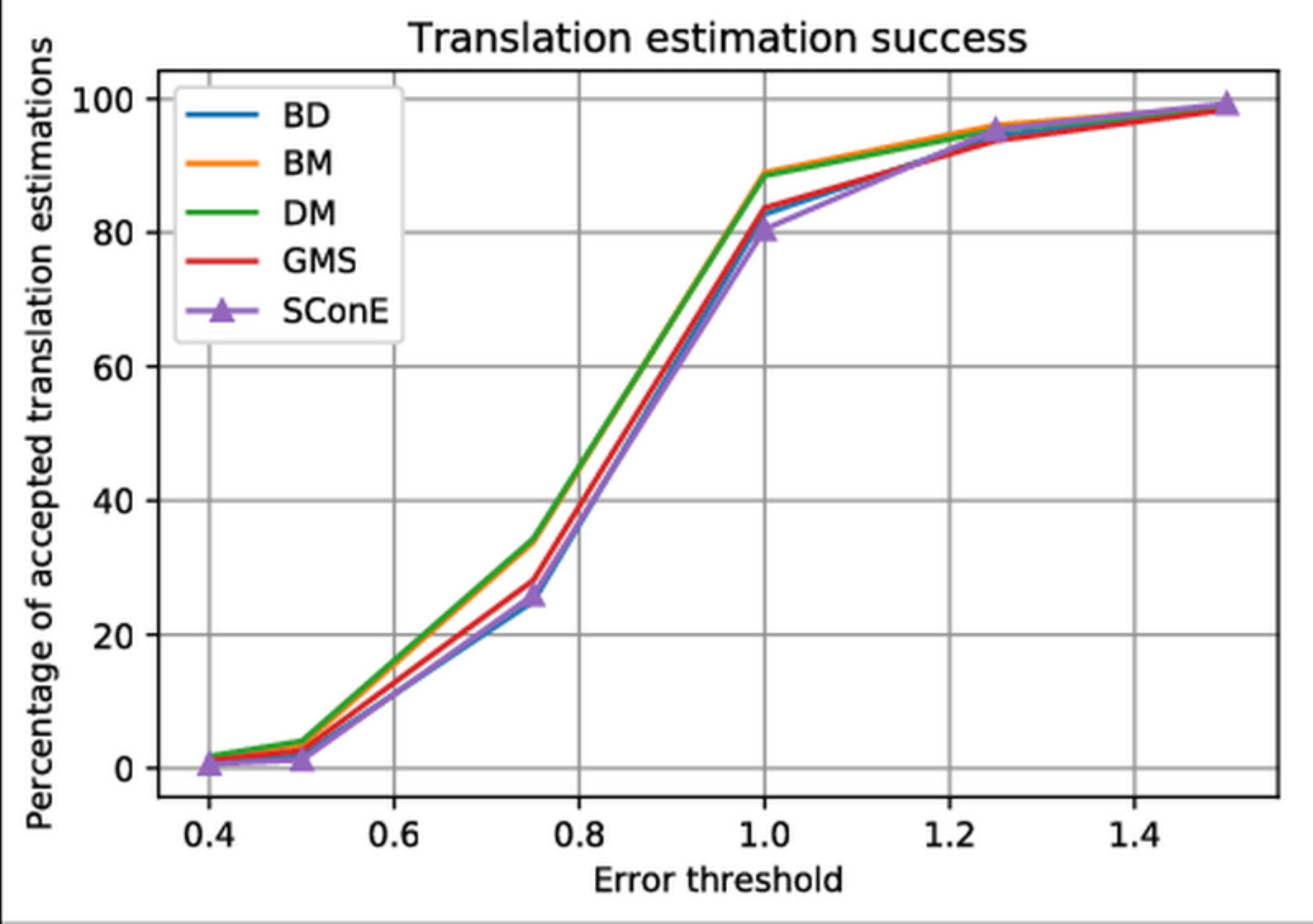}
\caption{Pose estimation errors on TUM dataset for SConE and significantly more complex matching procedures. Performance of SConE is au pair with much more advanced keypoint matching procedures.}
\label{fig:sfm_results1}
\end{figure}

In addition to GMS and basic matchers, we use DeepMatching (DM \cite{weinzaepfel2013dm}), Bilateral Functions Matching (BM \cite{wenyan2014bf}) and Bounded Distortion (BD \cite{Lipman2014bd}) as state-of-the-art image matchers. We use original implementations of its authors, so its computational efficiency may be considered as far from optimal. Where possible, we use only one computing thread for better comparison. 

\subsection{Results}
\label{sec:results}

We analyse our results using pose estimation errors obtained for various descriptors. Fig.~\ref{fig:sfm_results2} shows the results of this experiment. SConE outperforms all basic features with ratio tests. The performance gap is substantial, and shows gain over basic FREAK descriptor. SConE uses FREAK keypoints as a base descriptor for learning, thus it has the same keypoints pool before filtering stage. This characteristic lets us build simple comparison based solely on true positives after ratio test. Fig.~\ref{fig:gt1} shows number of true positives on FREAK keypoint locations, using both FREAK descriptors and SConE embeddings. The X-axis contains absolute distance between frames calculated as the difference between quaternion rotations for each camera position. 

SConE gives very good results in comparison with advanced matching methods, as shown in Fig.~\ref{fig:sfm_results1}. All of the descriptors give very similar results in translation estimation. Rotation estimation is much more prone to keypoint localization perturbations, thus shows more variance between methods. Our approach outperforms GMS with its default keypoints pool (10 000 ORB keypoints).

\begin{figure}
\centering
\includegraphics[width=0.5\textwidth]{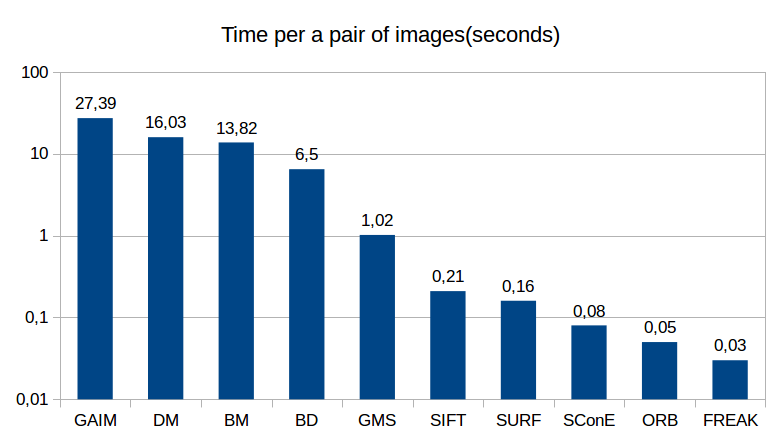}
\caption{Descriptor extraction and matching time for matching descriptors between a pair of images.
SConE offers very competitive performance compared to more sophisticated matching methods.}
\label{fig:running_times}
\end{figure}

Furthermore, we evaluate the computational complexity of our proposed SConE descriptor-based matching and compare it with the competing methods. 
We measure both descriptor extraction and matching times. For raw descriptors we use brute force matcher. Fig.~\ref{fig:running_times} shows the results of this comparison. SConE adds very little overhead to FREAK computation, which is used as base. It's much faster then GMS, while obtaining better results.

\section{Conclusions}
\label{sec:conclusions}
In this paper, we propose a novel low-dimensional feature descriptor that incorporates geometrical information about the layout of neighbouring keypoints. This way we are able to reduce the importance of additional post-processing step that typically aims at filtering out incorrect matches. To train our descriptor we use Siamese neural network architecture and feed it with central keypoint descriptor, as well as neighbouring keypoints and their descriptors, relative position, orientation and scale. Although our framework is agnostic to descriptor type, we use as our base descriptor FREAK and show that the SConE descriptor generated by our neural network outperforms competitors on a challenging TUM dataset. 

\section*{Acknowledgement}
This research was supported by Google Sponsor Research Agreement under
the project "Efficient visual localization on mobile devices".

The Titan X Pascal used for this research was donated by the NVIDIA Corporation.


\bibliographystyle{splncs}
\bibliography{egbib}

\begin{thebibliography}{10}

\bibitem{Agarwal2011}
Agarwal, S., Furukawa, Y., Snavely, N., Simon, I., Curless, B., Seitz, S.M.,
  Szeliski, R.:
\newblock Building rome in a day.
\newblock Commun. ACM \textbf{54}(10) (October 2011)  105--112

\bibitem{Brown2007}
Brown, M., Lowe, D.G.:
\newblock Automatic panoramic image stitching using invariant features.
\newblock IJCV \textbf{74}(1) (2007)  59--73

\bibitem{Lynen15}
Lynen, S., Sattler, T., Bosse, M., Hesch, J.A., Pollefeys, M., Siegwart, R.:
\newblock Get out of my lab: Large-scale, real-time visual-inertial
  localization.
\newblock In: Robotics: Science and Systems. (2015)

\bibitem{lowe2004distinctive}
Lowe, D.G.:
\newblock Distinctive image features from scale-invariant keypoints.
\newblock IJCV \textbf{60}(2) (2004)  91--110

\bibitem{alahi2012freak}
Alahi, A., Ortiz, R., Vandergheynst, P.:
\newblock Freak: Fast retina keypoint.
\newblock In: CVPR. (2012)

\bibitem{bay2006surf}
Bay, H., Tuytelaars, T., Van~Gool, L.:
\newblock Surf: Speeded up robust features.
\newblock In: ECCV. (2006)

\bibitem{rublee2011orb}
Rublee, E., Rabaud, V., Konolige, K., Bradski, G.:
\newblock Orb: An efficient alternative to sift or surf.
\newblock In: ICCV. (2011)

\bibitem{Trzcinski13a}
Trzcinski, T., Christoudias, M., Lepetit, V., Fua, P.:
\newblock {Boosting Binary Keypoint Descriptors}.
\newblock In: CVPR. (2013)

\bibitem{Simo_iccv2015}
Simo-Serra, E., Trulls, E., Ferraz, L., Kokkinos, I., Fua, P., Moreno-Noguer,
  F.:
\newblock Discriminative learning of deep convolutional feature point
  descriptors.
\newblock In: ICCV. (2015)

\bibitem{yi2016lift}
Yi, K.M., Trulls, E., Lepetit, V., Fua, P.:
\newblock Lift: Learned invariant feature transform.
\newblock In: ECCV. (2016)

\bibitem{sturm2012benchmark}
Sturm, J., Engelhard, N., Endres, F., Burgard, W., Cremers, D.:
\newblock A benchmark for the evaluation of rgb-d slam systems.
\newblock In: IROS. (2012)

\bibitem{han2015matchnet}
Han, X., Leung, T., Jia, Y., Sukthankar, R., Berg, A.C.:
\newblock Matchnet: Unifying feature and metric learning for patch-based
  matching.
\newblock In: CVPR. (2015)

\bibitem{loquercio2017efficient}
Loquercio, A., Dymczyk, M., Zeisl, B., Lynen, S., Gilitschenski, I., Siegwart,
  R.:
\newblock Efficient descriptor learning for large scale localization.
\newblock In: ICRA. (2017)

\bibitem{Niepert:2016:LCN:3045390.3045603}
Niepert, M., Ahmed, M., Kutzkov, K.:
\newblock Learning convolutional neural networks for graphs.
\newblock In: Proceedings of the 33rd International Conference on International
  Conference on Machine Learning - Volume 48. ICML'16, JMLR.org (2016)
  2014--2023

\bibitem{Defferrard:2016:CNN:3157382.3157527}
Defferrard, M., Bresson, X., Vandergheynst, P.:
\newblock Convolutional neural networks on graphs with fast localized spectral
  filtering.
\newblock In: Proceedings of the 30th International Conference on Neural
  Information Processing Systems. NIPS'16, USA, Curran Associates Inc. (2016)
  3844--3852

\bibitem{muja2012fast}
Muja, M., Lowe, D.G.:
\newblock Fast matching of binary features.
\newblock In: Computer and Robot Vision (CRV). (2012)

\bibitem{muja2014scalable}
Muja, M., Lowe, D.G.:
\newblock Scalable nearest neighbor algorithms for high dimensional data.
\newblock TPAMI \textbf{36}(11) (2014)  2227--2240

\bibitem{fischler1981random}
Fischler, M.A., Bolles, R.C.:
\newblock Random sample consensus: a paradigm for model fitting with
  applications to image analysis and automated cartography.
\newblock Communications of the ACM \textbf{24}(6) (1981)  381--395

\bibitem{raguram2013usac}
Raguram, R., Chum, O., Pollefeys, M., Matas, J., Frahm, J.M.:
\newblock Usac: a universal framework for random sample consensus.
\newblock TPAMI \textbf{35}(8) (2013)  2022--2038

\bibitem{bian2017gms}
Bian, J., Lin, W.Y., Matsushita, Y., Yeung, S.K., Nguyen, T.D., Cheng, M.M.:
\newblock Gms: Grid-based motion statistics for fast, ultra-robust feature
  correspondence.
\newblock In: CVPR. (2017)

\bibitem{Bromley:1993:SVU:2987189.2987282}
Bromley, J., Guyon, I., LeCun, Y., S\"{a}ckinger, E., Shah, R.:
\newblock Signature verification using a "siamese" time delay neural network.
\newblock In: NIPS. (1993)

\bibitem{Hadsell:2006:DRL:1153171.1153654}
Hadsell, R., Chopra, S., LeCun, Y.:
\newblock Dimensionality reduction by learning an invariant mapping.
\newblock In: CVPR. (2006)

\bibitem{NIPS2017_6698}
Klambauer, G., Unterthiner, T., Mayr, A., Hochreiter, S.:
\newblock Self-normalizing neural networks.
\newblock In Guyon, I., Luxburg, U.V., Bengio, S., Wallach, H., Fergus, R.,
  Vishwanathan, S., Garnett, R., eds.: Advances in Neural Information
  Processing Systems 30.
\newblock Curran Associates, Inc. (2017)  971--980

\bibitem{hochreiter1997long}
Hochreiter, S., Schmidhuber, J.:
\newblock Long short-term memory.
\newblock Neural computation \textbf{9}(8) (1997)  1735--1780

\bibitem{OpenCV}

\newblock https://opencv.org/

\bibitem{rosten2006machine}
Rosten, E., Drummond, T.:
\newblock Machine learning for high-speed corner detection.
\newblock In: ECCV. (2006)

\bibitem{Geiger2012kitty}
Geiger, A., Lenz, P., Urtasun, R.:
\newblock Are we ready for autonomous driving? the kitti vision benchmark
  suite.
\newblock In: CVPR). (2012)

\bibitem{weinzaepfel2013dm}
Weinzaepfel, P., Revaud, J., Harchaoui, Z., Schmid, C.:
\newblock {DeepFlow: Large displacement optical flow with deep matching}.
\newblock In: ICCV. (2013)

\bibitem{wenyan2014bf}
Lin, W.Y., Cheng, M.M., Lu, J., Yang, H., Do, M.N., Torr, P.:
\newblock Bilateral functions for global motion modeling.
\newblock In: ECCV. (2014)

\bibitem{Lipman2014bd}
Lipman, Y., Yagev, S., Poranne, R., Jacobs, D.W., Basri, R.:
\newblock Feature matching with bounded distortion.
\newblock ACM Trans. Graph. \textbf{33}(3) (June 2014)  26:1--26:14

\end{thebibliography}

\end{document}